\documentclass{article} % For LaTeX2e
\usepackage{tencent_tech_report}
\usepackage[colorlinks = true,
            linkcolor = blue,
            urlcolor  = blue,
            citecolor = blue,
            anchorcolor = blue]{hyperref}

\usepackage{tcolorbox} 
\usepackage{microtype}
\usepackage{hyperref}
\usepackage{url}
\usepackage{booktabs}
\usepackage{enumitem}
\usepackage{multicol}
\usepackage{multirow}
\usepackage{CJKutf8}
\usepackage{amsmath}
\usepackage{siunitx}
\usepackage{floatflt}
\usepackage{wrapfig}
\usepackage{graphicx}
\usepackage{booktabs}
\usepackage{wrapfig}
\usepackage{authblk}
\usepackage{lipsum}

\usepackage{algorithm}
\usepackage{algorithmicx}
\usepackage{algpseudocode}
\usepackage{microtype}
\usepackage{graphicx}
\usepackage{subfigure}
\usepackage{multirow}
\usepackage{booktabs} 
\usepackage{pifont}  
\usepackage{graphicx} 
\usepackage{subcaption}
\usepackage{hyperref}
\usepackage{amssymb}

\algnewcommand{\LeftComment}[1]{\Statex \(\triangleright\) #1}

\usepackage{array}
\usepackage{amsmath}
\usepackage{amssymb}
\usepackage{mathtools}
\usepackage{amsthm}
\usepackage{arydshln}
\usepackage[capitalize,noabbrev]{cleveref}
\usepackage{adjustbox}
\usepackage{enumitem}

\usepackage{listings}          
\tcbuselibrary{listings,skins}

\newtcolorbox{promptspecial}[1][]{
  colback      = gray!10,        
  colframe     = gray!50!black,  
  fontupper    = \ttfamily,      
  nobeforeafter,
  title        = Prompt Template,
  #1
}

\newtcolorbox{prompttext}[1][]{
  colback=gray!10,        
  colframe=gray!50!black, 
  nobeforeafter,
  title=Prompt Template,
  #1
}

\newtcolorbox{casestudy}[1][]{
  colback=gray!10,        % Light grey background
  colframe=gray!50!black, % Grey frame
  nobeforeafter,
  title=Case Study,
  #1 % Optional settings
}

\definecolor{nred}{RGB}{196, 38, 11}
\definecolor{ngreen}{RGB}{18, 141, 21}
\definecolor{nblue}{RGB}{41, 52, 190}

\newcommand{\method}{RISE}

\colmfinalcopy

\title{{\em Trust, But Verify:} A Self-Verification Approach to Reinforcement Learning with Verifiable Rewards}

\author[ ]{Xiaoyuan Liu\thanks{Work done when Xiaoyuan Liu, Zhiwei He, and Wenxuan Wang were interning at Tencent.}~~$^{,1,2}$}
\author[ ]{Tian Liang$^{1}$}
\author[ ]{Zhiwei He$^{1}$}
\author[ ]{Jiahao Xu$^{1}$}
\author[ ]{Wenxuan Wang$^{1}$}
\author[ ]{\\Pinjia He$^{\dag, 2}$}
\author[ ]{\mbox{Zhaopeng Tu}\thanks{Correspondence to: Zhaopeng Tu \textless zptu@tencent.com\textgreater~and Pinjia He \textless hepinjia@cuhk.edu.cn\textgreater.}~~$^{,1}$}
\author[ ]{\mbox{Haitao Mi}$^{1}$}
\author[ ]{Dong Yu$^{1}$}

\affil[1]{Tencent}
\affil[2]{The Chinese University of Hong Kong, Shenzhen}

\begin{document}

\maketitle

\begin{figure}[h]
\centering
\includegraphics[width=\columnwidth]{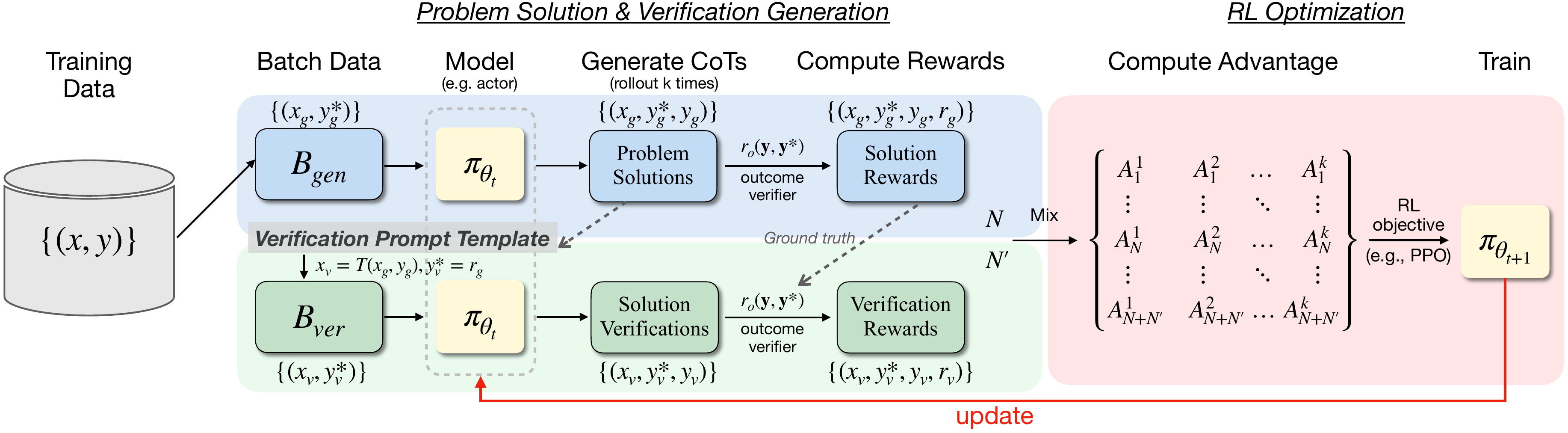}
%\vspace{-10pt}
\caption{Illustration of \method{}, which consists of two stages: (i) \textit{Problem Solution and Verification Generation}: problems from the training batch are used to generate chain-of-thought solutions from the model. Problems and model solutions are then formatted as verification prompts to generate verifications of the solutions. (ii) \textit{RL Optimization}: the original generation data and their verification are mixed as the new batch, and the model is optimized based on the RL objective.}
\label{fig1:key_method}
\end{figure}

\begin{abstract}
Large Language Models (LLMs) show great promise in complex reasoning, with Reinforcement Learning with Verifiable Rewards (RLVR) being a key enhancement strategy. However, a prevalent issue is ``superficial self-reflection'', where models fail to robustly verify their own outputs. We introduce \method{} (\textbf{R}einforcing Reason\textbf{i}ng with \textbf{S}elf-V\textbf{e}rification), a novel online RL framework designed to tackle this. \method{} explicitly and simultaneously trains an LLM to improve both its problem-solving and self-verification abilities within a single, integrated RL process. The core mechanism involves leveraging verifiable rewards from an outcome verifier to provide on-the-fly feedback for both solution generation and self-verification tasks. In each iteration, the model generates solutions, then critiques its own on-policy generated solutions, with both trajectories contributing to the policy update. 
Extensive experiments on diverse mathematical reasoning benchmarks show that \method{} consistently improves model's problem-solving accuracy while concurrently fostering strong self-verification skills. Our analyses highlight the advantages of online verification and the benefits of increased verification compute. Additionally, \method{} models exhibit more frequent and accurate self-verification behaviors during reasoning.
These advantages reinforce \method{} as a flexible and effective path towards developing more robust and self-aware reasoners.
\end{abstract}

\section{Introduction}

Large Language Models (LLMs) have demonstrated remarkable potential in complex reasoning tasks. A promising avenue for further enhancing these capabilities is Reinforcement Learning (RL), particularly methods that utilize verifiable rewards (RLVR) from outcome verifiers \citep{Gao2024OnDE, deepseekai2025deepseekr1incentivizingreasoningcapability, Lambert2024TLU3P, yue2025rlvrreally}. This paradigm, often applied to domains like mathematics where solution correctness can be programmatically evaluated, enabling models to improve through direct feedback on their generated solutions.

However, even with outcome-based RL, models may learn to generate correct answers without deeply understanding the underlying reasoning process or robust self-assessment skills. This can lead to ``superficial self-reflection''~\citep{liu2025oatzero}, where models struggle to reliably identify flaws in their own reasoning and verify the correctness of their outputs. While some approaches explicitly incorporate self-critique~\citep{xi2024enhancing, xie2025teaching}  to provide additional signals, the process of learning to solve problems and learning to verify solutions are often decoupled or lack direct, contemporaneous feedback for the verification skill itself within the RL loop.

To address this limitation and foster more robust reasoning, we introduce \textbf{\method{}} ({\bf R}einforcing Reason{\bf i}ng with {\bf S}elf-V{\bf e}rification) as a novel online reinforcement learning framework. \method{} is designed to explicitly and simultaneously train an LLM to improve both its problem-solving ability and its capacity to verify its own generated solutions within a single, integrated RL process. The key idea is to leverage the verifiable reward signal from a rule-based outcome verifier not only to guide the generation of correct solutions but also to align the model's self-verification ability on-the-fly.

In the \method{} framework, during each training iteration, the model first generates solutions for a batch of problems. Subsequently, using these on-policy generated solutions and the original problems, verification problems are constructed with a predefined template, prompting the model to critique its own solution and provide a score. The same outcome verifier used to assess problem solutions also provides ground-truth supervision for the verification task, based on an exact match between the predicted verification score and the ground-truth solution score. Both the problem-solving trajectories and the self-verification trajectories, along with their respective verifiable rewards, are then combined to update the model's parameters using a unified RL objective. This tight coupling enables the model to learn not only to solve problems, but also to critique and verify its own outputs, fostering a more dynamic and grounded self-improvement loop.

In our experiments, we implement and evaluate \method{} using the Proximal Policy Optimization (PPO) algorithm, applying it to the 1.5B, 3B, and 7B base models from the Qwen2.5 series.
Compared to a Zero-RL baseline, which incorporates only problem-solving supervision, \method{} consistently improves reasoning accuracy and achieves up to a $2.8\times$ increase in verification accuracy on challenging mathematical benchmarks. Moreover, \method{} outperforms instruction-tuned models across both tasks. For instance, \method{}-3B achieves a 3.7\% average improvement in reasoning accuracy and a 33.4\% gain in self-verification accuracy over Qwen-3B-Instruct.

We also find that this enhanced self-verification ability contributes to improved test-time performance. Specifically, \method{}-3B and \method{}-7B outperform standard majority voting by +0.2\% and +1.9\%, respectively, under a $k{=}4$ inference budget. Further analysis reveals that \method{} enhances the internal reasoning process by encouraging more frequent and effective verification behaviors. Finally, our ablations demonstrate that online verification is crucial to the success of \method{}.

Our main contributions are as follows:
\begin{itemize}[leftmargin=12pt]
% 2025-05-14 zptu
   \item We introduce \textbf{\method{}} (\textbf{R}einforcing Reason\textbf{i}ng with \textbf{S}elf-V\textbf{e}rification), a novel online reinforcement learning framework that explicitly and simultaneously trains LLMs to improve both problem-solving and self-verification capabilities within a single, integrated RL process, leveraging verifiable rewards for both tasks on-the-fly.
   \item We demonstrate, through extensive experiments on challenging mathematical reasoning benchmarks using a PPO-based implementation, that \method{} significantly boosts problem-solving performance while instilling robust self-verification skills in the LLM.
    \item We provide comprehensive analyses elucidating the critical role of \method{}'s online verification mechanism, the benefits of scaling verification training compute, and how the developed self-verification capability contributes to more accurate and reliable solution generation.
\end{itemize}

\section{Related Work}
\paragraph{RLVR for LLM Reasoning}
In the literature, reinforcement learning has been widely used to align language models with human preferences, typically through reward models or pairwise preference comparisons~\citep{christiano2017deep, ouyang2022training, rafailov2024direct}. More Recently, Reinforcement Learning with Verifiable Rewards (RLVR) has emerged as a powerful approach for improving the reasoning capabilities of LLMs in domains such as mathematics and programming~\citep{jaech2024openai, deepseekai2025deepseekr1incentivizingreasoningcapability}. Using only outcome rewards, recent work has demonstrated the scalability of RL algorithms for LLM reasoning~\citep{deepseekai2025deepseekr1incentivizingreasoningcapability,team2025kimi, zeng2025simplerlzoo, hu2025open}. However, leveraging verifiable rewards not only for reasoning supervision but also as a direct training signal for self-verification remains underexplored, which is the main focus of \method{}.

\paragraph{Learning to Solve and Verify}
Solution generation and verification are two foundational capabilities of LLMs~\citep{huang2024sharpening, song2024mind}, echoing the classic P versus NP dichotomy in computer science~\citep{enwiki:pnp}. In the context of LLM reasoning, previous work has focused on teaching models either to solve problems~\citep{deepseekai2025deepseekr1incentivizingreasoningcapability, zelikman2022star}, to verify solutions~\citep{Wang2023MathShepherdVA, lightman2023let, shi2025heimdall, zhang2025generativever}, or to leverage the verification capability to perform for self-improvement~\citep{Yuan2024SelfRewardingLM, xiong2025self}. More recently, \citet{lin2025learning} proposed a self-play framework that jointly teaches LLMs to generate code and corresponding test cases through two-stage training. In contrast, we introduce an online RL framework that explicitly leverages verifiable reward signals to jointly align the model's problem-solving and self-verification abilities in a unified training process.

\section{Reinforcement Learning Preliminaries}

\paragraph{Policy Gradient Methods}
The goal of RL is to learn a policy that maximizes the expected cumulative reward (namely return), denoted as the performance measure $J$.
Policy gradient methods learn a parameterized policy that can select actions to maximize $J$ without consulting other value functions. Grounded by the \textit{policy gradient theorem}~\citep{sutton2018reinforcement}, the optimization is performed as gradient ascent based on the gradient of $J(\theta)$ with respect to the policy parameter $\theta$. 

A large language model is naturally a parametrized policy $\pi_\theta$. The state at time~$t$, denoted as $s_t$, is the concatenation of the prompt~$\mathbf{x}$ and the response~$\mathbf{y}_{<t}$ generated so far, while the action $a_t$ is the next token~$y_t$. $T$ refers to total timestamps (response length + 1). Thus, the gradient can be expressed as:
\begin{equation*}
    \nabla_\theta J(\theta) = \mathbb{E}_{{\mathbf{x}\sim\mathcal{D},\,\mathbf{y}\sim\pi_\theta}}
    \!\left[\sum_{t=0}^{T} \nabla_\theta \log \pi_\theta(y_t \mid \mathbf{x}, \mathbf{y}_{<t})\,A_t\right].
\end{equation*}

The core part of this method is the advantage function $A_t$, which determines the extent to increase or decrease the probability of selecting this action (token) in the given state. In practice, the advantage function is implemented as cumulative discounted rewards subtracting an optional baseline, representing how much better an action is compared to the alternatives:
\begin{equation}
\label{eq:reinforce}
    A_t = \sum_{t=t_0}^{T} \gamma^{t-t_0} r_t - b(s_{t_0}),
\end{equation}
where $\gamma\!\in\![0,1]$ is the discount factor for the future rewards and $r_t=R(s_t, a_t, s_{t+1})$ is the reward from the environment at time~$t$.
Different implementations of the baseline formulate multiple variants of policy gradient methods, including using learned state-value functions (e.g., REINFORCE~\citep{williams1992simple}, Actor-Critic~\citep{actor-critic}), group-level reward means (e.g., GRPO~\citep{deepseek-math}), and leave-one-out (e.g. RLOO~\citep{ahmadian2024back}).

\paragraph{Proximal Policy Optimization}
Proximal Policy Optimization (PPO)~\citep{schulman2017proximal} is a popular algorithm of Actor-Critic method, which incorporates a critic model $\phi$ to help estimate advantage for training the actor model $\theta$ (i.e., policy). One major improvement of PPO is penalizing excessive policy updates and thereby maintaining training stability. In practice, the objective of the actor model is defined as follows:
\begin{equation}
% \begin{split}
    \mathcal{J}(\theta) = \mathbb{E}_{t} \left[ \min ( r_t(\theta) \hat{A}_t, \text{clip} \left(r_t(\theta), 1 - \epsilon, 1 + \epsilon )  \hat{A}_t \right) - \beta {KL}\left(\pi_{\theta} || \pi_{ref})\right) \right],
% \end{split}
\label{eq:actor_loss}
\end{equation}
where $r_t(\theta) = \frac{\pi_\theta(y_t |x, y_{<t})}{\pi_{\theta_{\text{old}}}(y_t |x, y_{<t})}$. Clip() and KL() are two techniques used for limiting update magnitudes. With Generalized Advantage Estimation (GAE)~\citep{SchulmanMLJA15}, the advantage is estimated as a $\lambda$-weighted sum of step-emporal-Difference (TD) errors:
\begin{equation}
\label{eq:gae}
    \hat{A}_t = \delta_t + (\gamma\lambda)\delta_{t + 1} + \cdots + (\gamma\lambda)^{T - t + 1}\delta_{T - 1},
\end{equation}
\begin{equation*}
    \text{where } \delta_t = r_t + \gamma V_\phi(s_{t + 1}) - V_\phi(s_t).
\end{equation*}
$T$ denotes response length with token indexes from $0$ to $T - 1$. $V_\phi(s_t)$ is the value predicted by the critic model $\phi$ at state $s_t$, $r_t$ is the scalar reward from the environment at time $t$, and $\lambda \in [0,1]$ is the GAE parameter that trades off between bias and variance. 
In practice, we set $\lambda=\gamma=1$, thus making the per-token loss averaged over the full response length $T$. By design, $r_t = 0$ for $t < T - 1$, and $r_t = r$ for $t = T - 1$ (i.e., outcome reward). 
After we update the actor model, the critic model should also be updated for accurate value estimations. In practice, we use Mean Squared Error (MSE) to measure the prediction loss and perform the update:
\begin{equation}
\label{eq:critic_loss}
\mathcal{J}(\phi) =  \mathbb{E}_t \left[ \max\left( 
(V_{\phi}(s_t) - V_t^{targ})^2,\ 
\left(\text{clip}(V_{\phi}(s_t), V_{\phi_\text{old}}(s_t) - \epsilon, V_{\phi_\text{old}}(s_t) + \epsilon) - V_t^{targ} \right)^2 
\right) \right],
\end{equation}
\begin{equation*}
    \text{where } V_t^{targ} = V_{\phi_\text{old}}(s_{t}) + \hat{A}_t.
\end{equation*}

\paragraph{Verifiable Reward}
Unlike the rewards from conventional reward models which are continuous numbers denoting the goodness of the response, verifiable rewards are usually discrete numbers representing the correctness of the final result~\citep{Lambert2024TLU3P,deepseekai2025deepseekr1incentivizingreasoningcapability}. Given the prompt $\mathbf{x}$ and the complete response $\mathbf{y}$ from the LLM $\pi_\theta$, the verifiable reward is given as a binary number by a deterministic outcome verifier $OV$: $r = OV(\mathbf{x}, \mathbf{y}) \in \{0, 1\}$, where $r = 1$ if and only if the final answer is exactly correct (e.g., the numeric result is mathematically identical to the ground truth answer) and $r = 0$ otherwise. In practice, an auxiliary format reward can be included to encourage the model to present its answer in a prescribed style.

\section{Methodology: Reinforcing Reasoning with Self-Verification (\method{})}

To address the challenge of superficial self-reflection, we propose \method{} for self-improving reasoners, which is a scalable online RL method with explicit verification objective. \textbf{The key idea of \method{} is the use of the verifiable reward signal from the rule-based outcome verifier to align the model's verification ability on-the-fly.} This enables us to teach the model to verify its own response at the same time it solves the problem, as depicted in Figure~\ref{fig1:key_method} and Algorithm~\ref{algo:rise}.

\subsection{Online Reasoning and Verification}
\label{sec:online_gen}
\paragraph{Problem Solution Generation}
Given an initial model $\pi_{\theta}$ and a training set $D = \{(\mathbf{x_i}, \mathbf{y}_i^*)\}$ consisting of problems $\mathbf{x}_i,$ and their corresponding ground-truth answers $\mathbf{y}_i^*$, we begin each RL iteration by sampling a data batch. At iteration $t$, the model first generates $k$ solutions for each problem in the batch, each comprising a chain-of-thought reasoning followed by a final answer.

Next, the reward is computed for each generated response. Following prior RLVR approaches, we define a rule-based outcome verifier (OV) that incorporates both answer and format correctness:
\begin{align*}
r_o(\mathbf{y}, \mathbf{y}^*) &= \begin{cases}1, &\text{boxed and matched} \\ -0.5, &\text{boxed but not matched}\\-1,&\text{unboxed} \end{cases} 
\end{align*}
Here ``matched'' means the final answer in the generated solution $\mathbf{y}$ is mathematically identical to the provided ground truth $\mathbf{y}^*$, and ``boxed'' means the final answer in $\mathbf{y}$ is wrapped in the \texttt{\textbackslash boxed\{\}}. 

This produces the generation batch $\mathcal{G} = {(\mathbf{x}, \mathbf{y}, r)}$, where each element includes the input problem, a model-generated solution, and its associated reward.

\paragraph{Online Solution Verification}
To construct verification data, we apply a fixed prompt template (Figure~\ref{fig:ver_prompt}) to $\mathcal{G}$, formatting the problem-solution pair into a new verification prompt $\mathbf{x}_\text{ver}$ that explicitly states the verification criteria and asks the model to critique the provided solution and assign a score. Since the criteria used in the prompt are exactly those employed by the outcome verifier, the original reward $r$ from the generation phase is reused as the ground-truth score for the verification task.
Thus, for each triple $(\mathbf{x}, \mathbf{y}, r) \in \mathcal{G}$, we construct the verification data as $(\mathbf{x}_\text{ver} = \mathcal{T}(\mathbf{x}, \mathbf{y}),\ \mathbf{y}^*_\text{ver} = r)$. In practice, the amount of verification data is controlled by the verification batch size.

For each verification prompt, the model generates $K$ responses, each containing a natural language critique and a final score. These responses are evaluated using the same OV criteria, where the reward is determined by whether the extracted score from the model’s response matches the ground-truth score. This process yields the verification batch $\mathcal{V} = {(\mathbf{x}, \mathbf{y}, r)}$, maintaining the same structure.

\subsection{RL Integration}
\label{sec:rl_optimize}
\begin{algorithm}[t]

\caption{\method{} (PPO)}
\label{algo:rise}

\textbf{Input} Language model $\pi_{\theta_{\text{init}}}$; outcome verifier OV; dataset $\mathcal{D}$; rollout number $K$; generation batch size $\mathcal{B}_g$, verification batch size $\mathcal{B}_v$; verification prompt template $\mathcal{T}$; total iteration $N$.

\begin{algorithmic}[1]
  \State \textbf{Initialize:} actor $\pi_{\theta}\!\leftarrow\pi_{\theta_{\text{init}}}$, old‑actor $\pi_{\theta_{\text{old}}}$, critic $\pi_{\phi}$, reference $\pi_{\text{ref}}$\vspace{2pt}

  \For{$\text{iteration}=1$ \textbf{to} $N$}
    \State Sample $\mathcal{B}_{g}$ samples for generation $\displaystyle \mathcal{P}_{g}=\{(\mathbf{x}_i,\mathbf{y}_i^*)\}_{i=1}^{\mathcal{B}_{g}}\sim\mathcal{D}$
    \State Get generation batch: \Comment{Generate solutions}
    \Statex \hspace{\algorithmicindent} $\displaystyle 
           \mathcal{G}\gets
           \bigl\{(\mathbf{x}_i,\mathbf{y}_{i}^{(k)},r_{\text{ov}}(\mathbf{y}_{i}^{(k)}, \mathbf{y}_i^*))
           \,\bigm|\,
           \mathbf{y}_{i}^{(k)}\!\sim\!\pi_\theta(\cdot|\mathbf{x}_i),
           \,i\!\le\!\mathcal{B}_g,\,k\!\le\!K\bigr\}$
    \State Select $\mathcal{B}_v$ triples $\mathcal{P}'=\{(\mathbf{x}_i,\mathbf{y}_i, r_i)\}_{i=1}^{\mathcal{B}_{v}}\subseteq\mathcal{G}$ for verification
    \State $\displaystyle
           \mathcal{P}_v\gets
           \bigl\{ \bigl(\mathcal{T}(\mathbf{x},\mathbf{y}),\,r\bigr)
           \mid (\mathbf{x},\mathbf{y},r)\in\mathcal{P}' \bigr\}
           \quad\text{\color{blue}{// each element is a new prob-ans tuple} $(\mathbf{x},\mathbf{y^*})$}$
    \State Get verification batch: \Comment{Verify generations}
    \Statex \hspace{\algorithmicindent} $\displaystyle
           \mathcal{V}\gets
           \bigl\{(\mathbf{x}_j,\mathbf{y}_{j}^{(k)},r_{\text{ov}}(\mathbf{y}_{j}^{(k)}, \mathbf{y}_j^*))
           \,\bigm|\,
           \mathbf{y}_{j}^{(k)}\!\sim\!\pi_\theta(\cdot|\mathbf{x}_j),
           \,j\!\le\!\mathcal{B}_v,\,k\!\le\!K\bigr\}$

    \State Get complete training batch $\mathcal{B}\leftarrow\mathcal{G}\cup\mathcal{V}$
    \State Estimate advantages $\hat{A}$ using Eq.~\eqref{eq:gae} \Comment{Joint optimization}
    \State Update critic $\pi_{\phi}$ by critic loss in Eq.~\eqref{eq:critic_loss}
    \State Update actor $\pi_{\theta}$ by actor loss in Eq.~\eqref{eq:actor_loss};\;update $\theta_{\text{old}}\leftarrow\theta$
  \EndFor
\end{algorithmic}
\textbf{Output} Optimized actor model $\pi_\theta$

\end{algorithm}

The preceding \emph{Online Reasoning and Verification} stage is architecturally agnostic to the choice of the underlying policy‑gradient algorithm; its only algorithm‑specific interface is the advantage estimator \(\hat A\) used in the policy update. In our formulation, advantage values are computed from a concatenated mini‑batch \(\mathcal B=\mathcal G\cup\mathcal V\), which aggregates rewards from both the generation and verification tasks. Since every sample in \(\mathcal B\) is annotated with a scalar reward and the action log‑probability under the current policy \(\pi_\theta\), any estimator that maps a sequence of state–action–reward tuples to an advantage can be incorporated without further structural change.

For our main experiments with PPO (see Algorithm~\ref{algo:rise}), we apply GAE (Eq.~\ref{eq:gae}) independently to each trajectory. The generation and verification trajectories are jointly processed within the same stochastic gradient descent (SGD) step, enabling the actor to be optimized with respect to both types of data. Meanwhile, the shared critic learns a unified value function across tasks. PPO's clipping mechanism further ensures that updates remain stable within a consistent trust region.

\section{Experiment}
\subsection{Experiment Setup}
\label{sec:exp_setup}
\textbf{Dataset} 
We follow the previous study~\citep{zeng2025simplerlzoo} to utilize MATH-Hard (Level 3–5)~\citep{hendrycks2021measuring} as our training set, which in total comprising 8,523 problems. This training set is used for all SFT baselines, Zero-RL baselines, and \method{} models.

\textbf{Models} 
\label{sec:model_train}
We conduct our RL training experiments on three Qwen2.5 models~\citep{qwen2.5} with different sizes (i.e., 1.5B, 3B, and 7B) for their strong reasoning capabilities.
The RL training of our models is based on the verl~\citep{sheng2024hybridflow} framework with a train batch size of 1024 and a mini-batch size of 128. We follow~\citep{zeng2025simplerlzoo} by setting the sampling temperature to 1.0 and rollout 8 responses for each problem. The \method{} models have a default verification batch size 128. We set the RL configurations same for \method{} models and Zero-RL models, ensuring a fair comparison.

\textbf{Benchmarks} 
We evaluate model performance on standard mathematical reasoning benchmarks: MATH500~\citep{hendrycks2021measuring, lightman2023let}, Minerva Math~\citep{lewkowycz2022solving}, OlympiadBench~\citep{he2024olympiadbench}, and competition-level benchmarks AIME 2024 and AMC 2023. Following~\citep{zeng2025simplerlzoo}, we generate 8 responses per problem using a sampling temperature of 1.0, and report Pass@1 accuracy~\citep{chen2021evaluating} as the evaluation metric. Reasoning correctness is based on exact match of the final answer, and verification correctness depends on exact match between the predicted verification score and the score from the outcome verifier.

\subsection{Experimental Results}

\begin{table*}[t]
\centering
\caption{Detailed results of \method{} and other baseline methods on various math benchmarks. Zero-RL models are trained with the setting with \method{} except the verification objective.}
\label{tab:zero_shot_results}
\resizebox{1.0\linewidth}{!}{
\setlength{\tabcolsep}{3pt}
\begin{tabular}{lrrrrrrrrrrrr}
\toprule
& \multicolumn{6}{c}{\textit{\textbf{Reasoning}}} & \multicolumn{6}{c}{\textit{\textbf{Self-Verification}}} \\
\cmidrule(lr){2-7} \cmidrule(lr){8-13}
\textbf{Model} & \textbf{\small MATH} & \textbf{\small AIME} & \textbf{\small AMC} & \textbf{\small Mine.} & \textbf{\small Olym.} & \textbf{Avg.} & \textbf{\small MATH} & \textbf{\small AIME} & \textbf{\small AMC} & \textbf{\small Mine.} & \textbf{\small Olym.} & \textbf{Avg.} \\
\midrule
GPT-4o & 79.0 & 13.3 & 55.0 & 50.0 & 42.5 & 48.0 & 83.4 & 33.3 & 67.5 & 50.4 & 54.4 & 57.8 \\ \midrule
\multicolumn{5}{l}{\em Qwen2.5-1.5B}\\
Base & 2.0 & 0.0 & 1.9 & 0.8 & 0.6 & 1.1 & 19.4 & 21.9 & 22.7 & 15.9 & 21.1 & 20.2 \\
Instruct & 37.5 & 0.8 & 19.4 & 8.3 & 11.7 & 15.5 & 48.8 & 22.1 & 36.5 & 36.9 & 29.6 & 34.8 \\
SFT & 10.1 & 0.0 & 4.1 & 1.8 & 2.0 & 3.6 & 19.0 & 5.8 & 12.3 & 10.5 & 10.9 & 11.7 \\
Zero-RL & 55.3 & 2.1 & 25.9 & \textbf{17.4} & 19.5 & 24.0 & 54.1 & 5.0 & 30.7 & 21.0 & 23.0 & 26.8 \\
\textbf{\method{}-1.5B} & \textbf{54.6} & \textbf{2.9} & \textbf{27.5} & 17.2 & \textbf{19.8} & \textbf{24.4} & \textbf{75.9} & \textbf{85.0} & \textbf{70.6} & \textbf{66.0} & \textbf{74.9} & \textbf{74.5} \\ \midrule
\multicolumn{5}{l}{\em Qwen2.5-3B}\\
Base & 32.7 & 1.3 & 15.3 & 10.3 & 10.7 & 14.1 & 39.5 & 13.6 & 22.5 & 29.9 & 21.2 & 25.3 \\
Instruct & 61.0 & 3.8 & 34.1 & 25.6 & 24.6 & 29.8 & 65.6 & 21.0 & 45.5 & 37.6 & 35.0 & 40.9 \\
SFT & 14.4 & 0.4 & 5.3 & 2.9 & 2.8 & 5.2 & 21.5 & 2.1 & 10.9 & 17.9 & 13.2 & 13.1 \\
Zero-RL & 64.2 & 6.7 & 37.5 & \textbf{27.4} & \textbf{26.6} & 32.5 & 64.9 & 13.0 & 39.7 & 30.3 & 31.2 & 35.8 \\
\textbf{\method{}-3B} & \textbf{64.3} & \textbf{7.9} & \textbf{42.5} & 26.2 & \textbf{26.6} & \textbf{33.5} & \textbf{81.0} & \textbf{86.3} & \textbf{74.4} & \textbf{56.1} & \textbf{73.6} & \textbf{74.3} \\ \midrule
\multicolumn{5}{l}{\em Qwen2.5-7B}\\
Base & 38.3 & 2.1 & 21.9 & 11.9 & 13.2 & 17.5 & 58.4 & 45.9 & 51.5 & 48.4 & 48.4 & 50.5 \\
Instruct & 73.8 & 10.0 & 50.6 & \textbf{35.9} & 35.8 & 41.2 & 77.2 & 26.3 & 57.0 & 40.2 & 45.2 & 49.2 \\
SFT & 28.7 & 0.8 & 13.8 & 6.2 & 7.2 & 11.3 & 40.5 & 36.6 & 47.4 & 39.2 & 36.1 & 40.0 \\
Zero-RL & 74.5 & 12.1 & 51.3 & 34.2 & \textbf{36.7} & 41.7 & 75.9 & 21.7 & 56.5 & 37.3 & 41.6 & 46.6 \\
\textbf{\method{}-7B} & \textbf{74.8} & \textbf{12.5} & \textbf{55.9} & 34.6 & \textbf{36.7} & \textbf{42.9} & \textbf{83.8} & \textbf{75.0} & \textbf{72.5} & \textbf{48.6} & \textbf{65.9} & \textbf{69.2} \\ \bottomrule
\end{tabular}
}
\end{table*}

Table~\ref{tab:zero_shot_results} presents the results of \method{} across model sizes and benchmarks.

\paragraph{\method{} significantly enhances self-verification capabilities while improving reasoning performance.} \method{} models consistently outperform their Zero-RL counterparts across both reasoning and self-verification tasks on all model sizes. The improvement in self-verification is particularly dramatic - \method-1.5B achieves 74.5\% average verification accuracy compared to just 26.8\% for Zero-RL, representing a 47.7 percentage point improvement. 
This demonstrates that our integrated approach successfully develops robust self-verification skills while simultaneously enhancing problem-solving capabilities. 
Notably, {\bf the verification improvements are particularly pronounced on the most challenging benchmarks} like AIME24 and OlympiadBench, suggesting that \method{} enables models to better recognize their limitations and errors on difficult problems.

\paragraph{Scaling model size improves reasoning performance while maintaining strong verification capabilities.} Scaling model size from 1.5B to 7B parameters consistently enhances reasoning performance across all benchmarks. Interestingly, the verification performance of \method{} models remains consistently high across model sizes, with all models achieving over 69\% average accuracy. 
The ability to maintain strong verification capabilities while scaling reasoning performance aligns with our contribution of developing a framework that simultaneously improves both critical capabilities.

\paragraph{\method{} models outperform standard SFT and base models by a substantial margin.} The results clearly demonstrate that \method{} models substantially outperform their SFT and base model counterparts. For instance, \method-7B achieves 42.9\% average reasoning accuracy compared to just 11.3\% for SFT-7B and 17.5\% for the base model. This dramatic improvement highlights the effectiveness of our integrated RL approach that incorporates both problem-solving and self-verification objectives.

\subsection{Test-Time Scaling with Self-verification}
\label{sec:tts}

\begin{figure*}[h]
\centering
    \subfigure[k=4]{
    \includegraphics[width=0.35\textwidth]{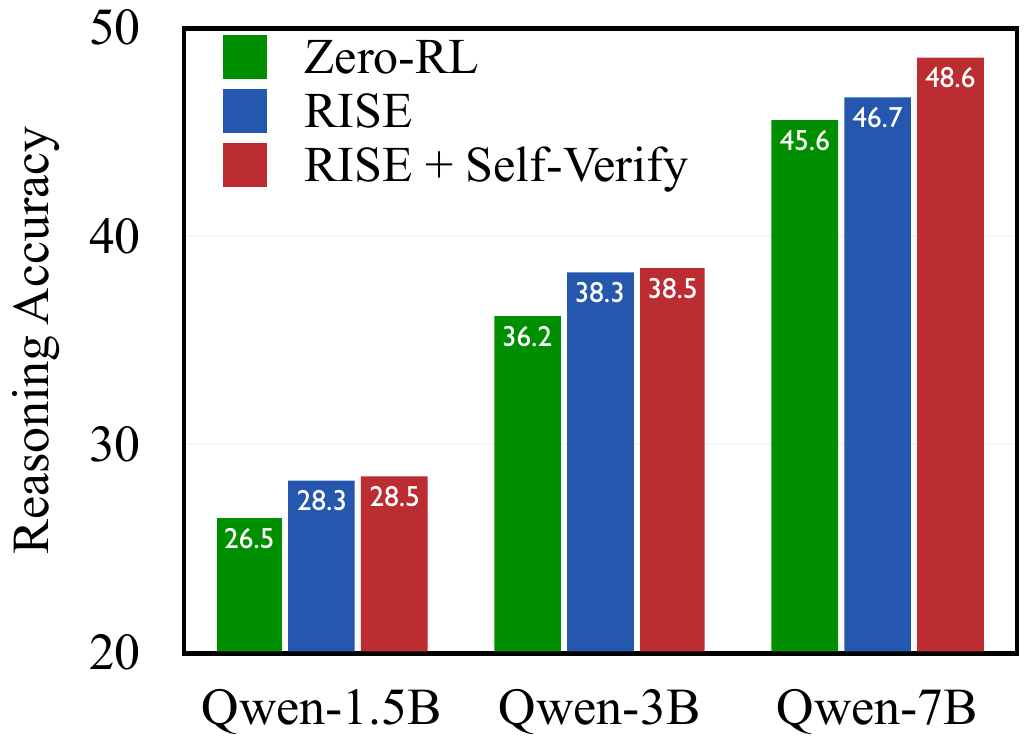}} \hspace{0.1\textwidth}
    \subfigure[k=8]{
    \includegraphics[width=0.35\textwidth]{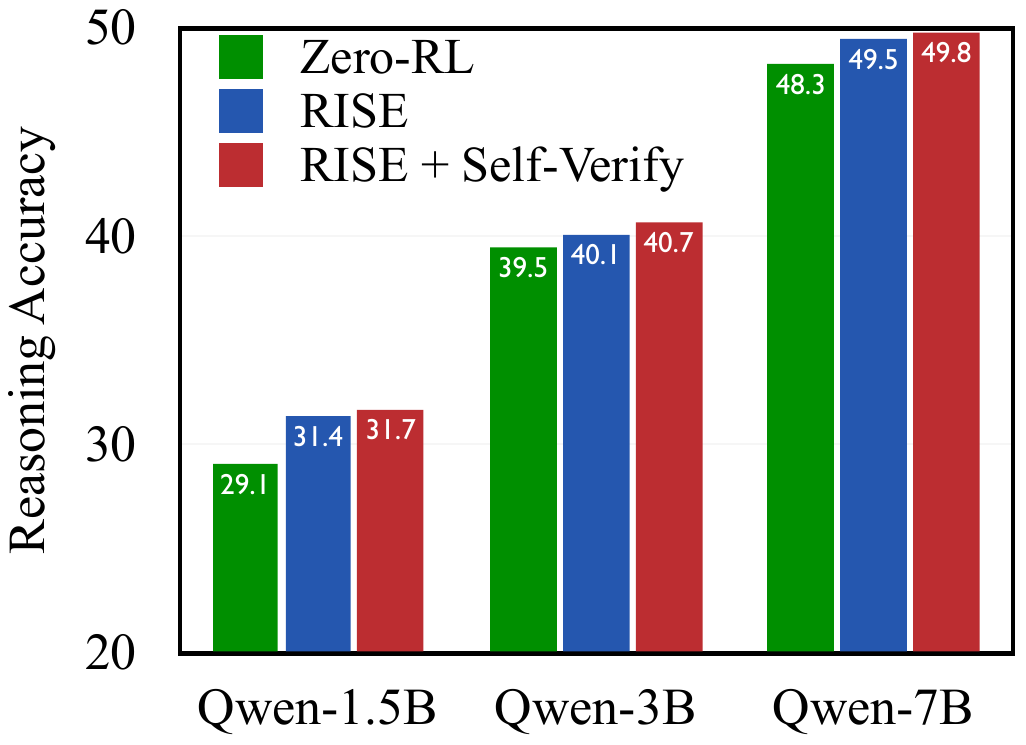}}
    \caption{Test-time scaling performance across different sampling budgets (``k'').}
    \label{fig:tts}
\end{figure*}

To further evaluate the benefits of the enhanced self-verification capabilities developed by \method{}, we investigate its impact at test-time using self-consistency majority voting (``maj@k'') \citep{wangself} and verification-weighted majority voting. In the latter, following \citep{Wang2023MathShepherdVA}, the model's self-generated verification scores for each candidate solution are used to weight its contribution in the majority vote. The results, presented in Figure \ref{fig:tts}, compare \method{} models against Zero-RL models across different sampling budgets (``k=4'' and ``k=8'').

\paragraph{\method{} consistently improves test-time scaling performance with self-verification and majority voting.} \method{} models outperform their Zero-RL counterparts when employing test-time strategies such as majority voting and verification-weighted selection. Across model sizes and sampling budgets, \method{} achieves higher average accuracy, with the largest relative gains observed when self-verification scores are used to re-rank majority votes. For example, \method{}-7B achieves an average score of 49.8\% with $k=8$ + self-verify, surpassing Zero-RL’s 48.3\% under the same conditions. This consistent improvement substantiates the effectiveness of integrating self-verification during both training and inference, fulfilling our objective of developing more robust and self-aware reasoners.

\paragraph{Verification-weighted voting delivers further accuracy gains.} Incorporating self-verification scores as weights in the voting process leads to additional accuracy improvements for all \method{} models. For instance, \method{}-3B and \method{}-7B models see improvements of +0.2\% and +1.9\% over standard majority voting at the $k=4$ budget, respectively. These results indicate that the self-verification policy learned by \method{} provides meaningful confidence signals for answer calibration.

\subsection{Comparison with Off-the-shelf Verifiers}
\label{para:ver_compare}

\begin{wrapfigure}{r}{0.4\textwidth}
  \centering
  \vspace{-25pt}
  \includegraphics[width=1.0\linewidth]{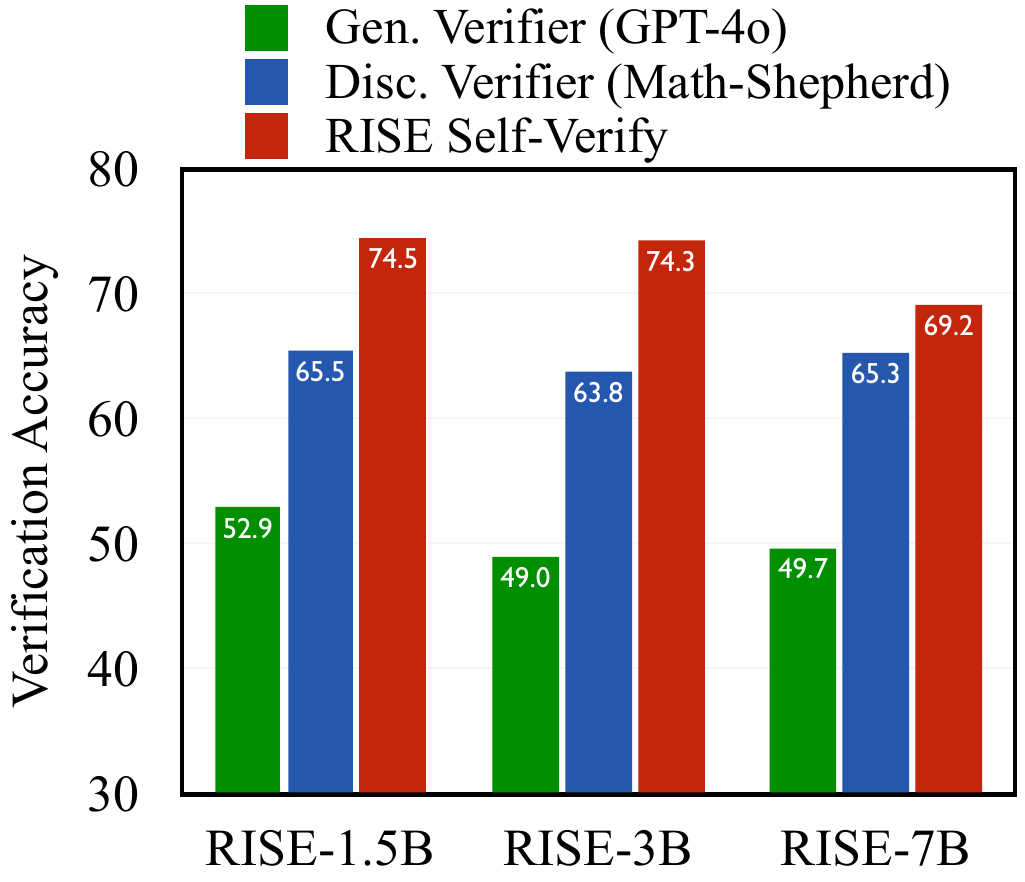} % Replace with your image
  \caption{Comparisons between \method{} (self-verify) and off-the-shelf verifiers.}
  \label{fig:v_acc_combined}
  \vspace{-10pt}
\end{wrapfigure}

We further compare the verification accuracy between our \method{} models as self-verifiers and off-the-shelf verifiers, including a discriminative verifier (Math-Shepherd-7B~\citep{Wang2023MathShepherdVA}) and a generative verifier (GPT-4o~\citep{gpt4o}). Specifically, we use the verification prompt in Figure~\ref{fig:ver_prompt} for both \method{} models and GPT-4o and adhere to the original logic for Math-Shepherd to verify the generated solutions. The results of \method{}-1.5B, 3B and 7B are presented in Figure~\ref{fig:v_acc_combined}, which show that \method{} models consistently outperform existing outcome verifiers in judge their solutions' correctness. This serves as a great advantage for the model to further improve its test-time performance, by leveraging the self-verification signal either externally or internally. Detailed results and evaluation implementation can be found in Appendix~\ref{apx:evaluation_details}.

\subsection{Analysis}

In this section, we provide some insights into how \method{} improves performance.

\begin{wrapfigure}{r}{0.6\textwidth}
  \centering
  \vspace{-10pt}
  \includegraphics[width=1.0\linewidth]{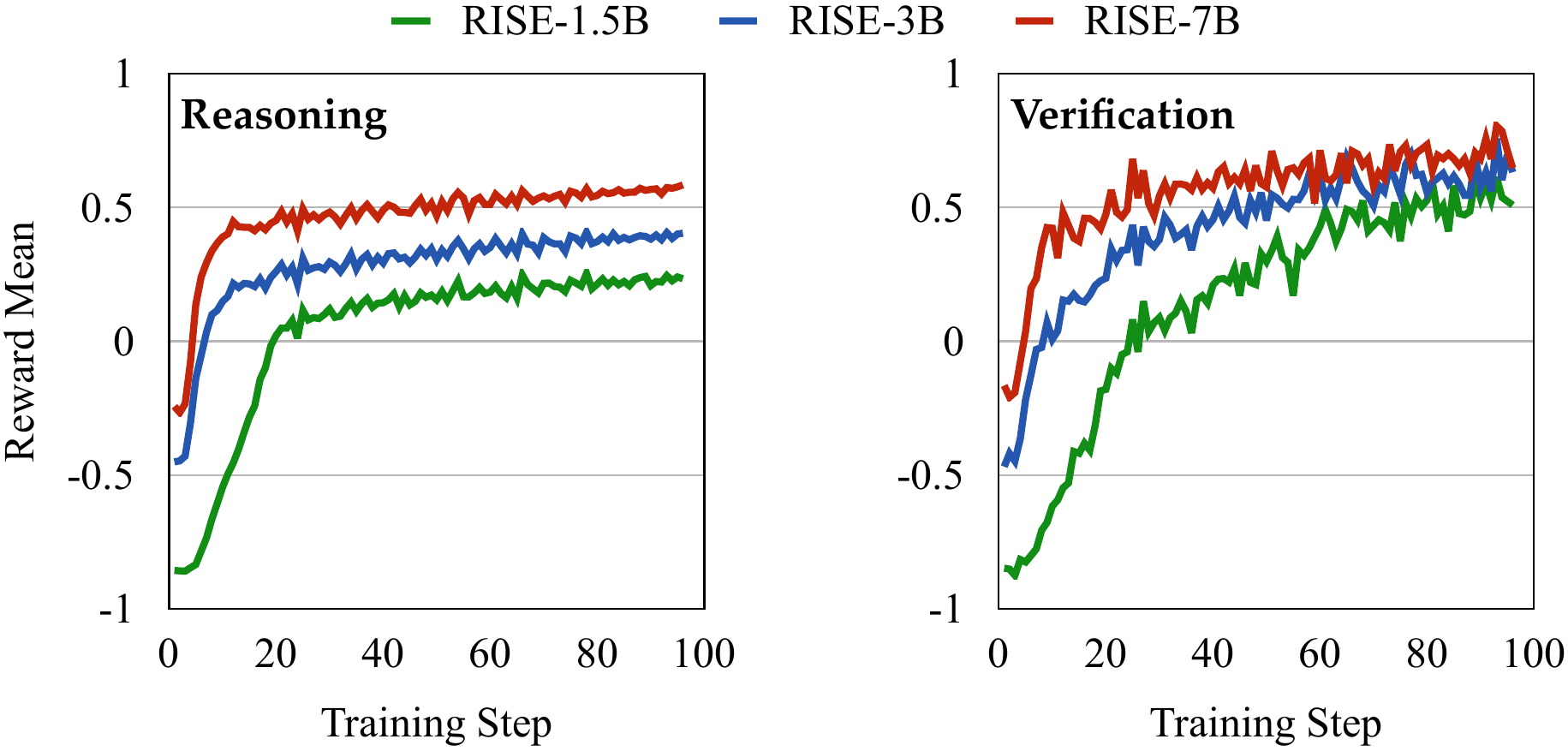} % Replace with your image
  \caption{Reasoning and verification reward at train time.}
  \label{fig:gen_ver_reward}
  % \vspace{-10pt}
\end{wrapfigure}
\paragraph{\method{} demonstrates robust and simultaneous learning of problem-solving and self-verification, with self-verification skills developing notably faster across different model scales.} The learning curves, illustrated by the reward trends in Figure~\ref{fig:gen_ver_reward}, reveal a consistent and steady improvement in both reasoning (problem-solving) and self-verification rewards throughout the RL training process for all evaluated models. This uniform positive progression across varying model sizes highlights the robustness of the \method{} framework in co-training these two abilities, a core contribution of our work.
A key observation is that the self-verification reward generally exhibits a more rapid increase and reaches a higher relative level compared to the problem-solving reward within the same training duration. This aligns with the ``Generation-Verification Gap'' posited by \citet{song2024mind}, suggesting that models might acquire verification capabilities more readily than complex reasoning.

\begin{wrapfigure}{r}{0.6\textwidth}
\centering
\vspace{-10pt}
\includegraphics[width=\linewidth]{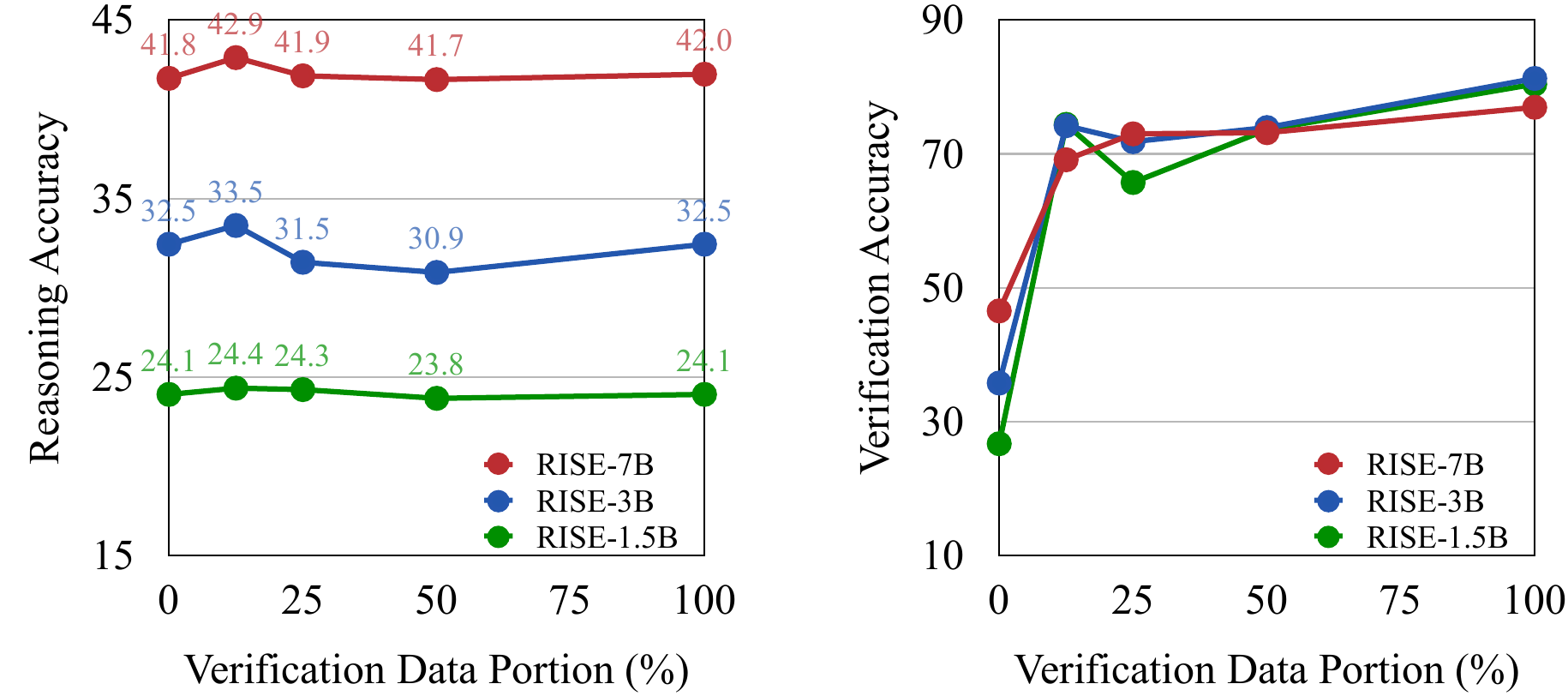}
\caption{Impact of verification data ratio.}
\label{fig:scaling_ver}
% \vspace{-10pt}
\end{wrapfigure}
\paragraph{Impact of Verification Compute}
\label{para:scaling_ver}
In the main experiment, we trained our \method{} models with a verification batch size of 128, which is 12.5\% of the generation batch 1024. We further explore the model performance by scaling up the verification data batch, i.e., the train-time compute, up to 100\% of the generation batch. In practice, we choose the percentages S of $\{0,12.5\%, 25\%, 50\%, 100\%\}$ and perform experiment on our \method{} models. The results are shown in Figure~\ref{fig:scaling_ver}. The problem-solving performance first increases, and then slightly decreases, and finally increases across the benchmark, maintaining at a high level. Furthermore, the verification performance keeps scaling with more training compute, indicating the robustness of scalability of our \method{} method.

\begin{wrapfigure}{r}{0.5\textwidth}
  \centering
  \vspace{-10pt}
  \includegraphics[width=1.0\linewidth]{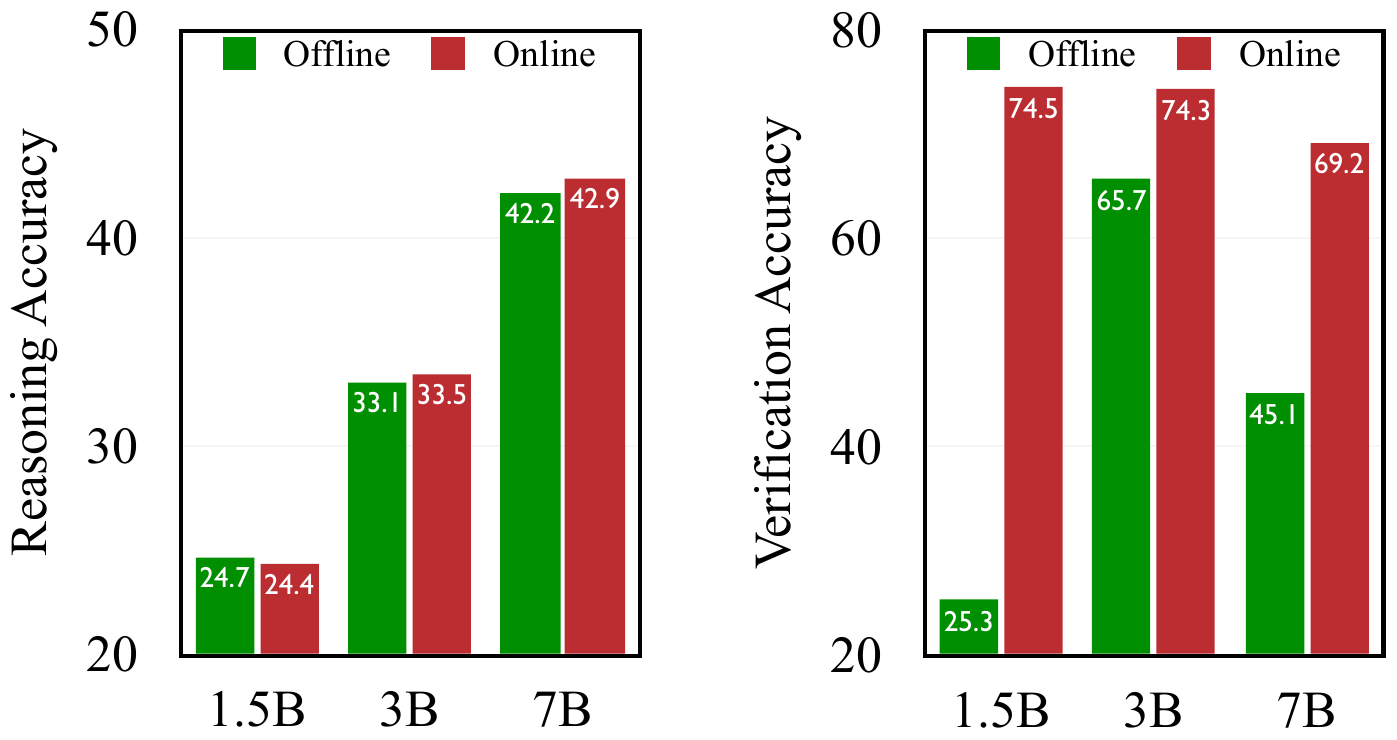} 
  \caption{Comparisons between online and offline verification.}
  \label{fig:online_offline}
  \vspace{-10pt}
\end{wrapfigure}

\paragraph{Online and Offline Verification}
We validate the effectiveness of online verification by comparing it to a offline variant, where the verification data are collected from a distant policy and directly added to the training set. In practice, we select the policy at step 96 (final step) of the Zero-RL model and use its generated responses to construct offline verification set. In the experiment, we keep the portion of verification data and the training batch size same to eliminate other influence factors, making the only changing variable the source of the verification data. 
Figure~\ref{fig:online_offline} shows the results. While the problem-solving performance of offline verification models are on par with the online ones, they have a significant drop in terms of self-verification accuracy, which indicates the importance of online verification designed in our \method{} method.

\begin{wrapfigure}{r}{0.5\textwidth}
  \centering
  \vspace{-10pt}
  \includegraphics[width=1.0\linewidth]{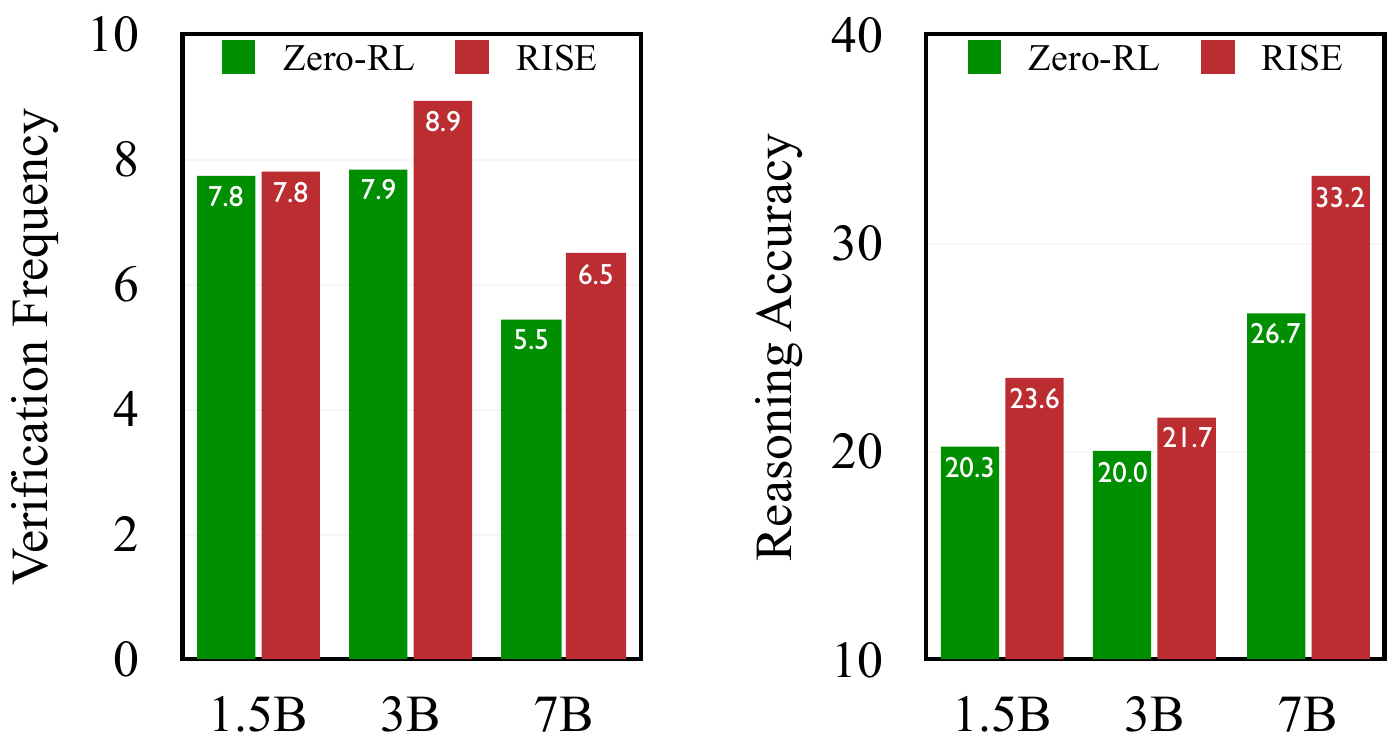} 
  \caption{Verification frequency (left panel) and its effectiveness in reasoning (right panel).}
  \label{fig:enhanced_verification}
  \vspace{-10pt}
\end{wrapfigure}

\paragraph{Enhanced Verification for Reasoning}
Besides leveraging the self-verification ability externally during the test-time as in \S~\ref{sec:tts}, such ability is also internalized by the model to enhance its reasoning generation process. To analysis this effect from the quantitative perspective, we measure the \emph{Verification Frequency} and \emph{Self-Verified Reasoning Accuracy} in models problem-solving process. Inspired by \citep{yeo2025demystifying}, we use a set of verification keywords to select the responses containing self-verification behaviors, namely \{``verify'', ``verifying'', ``recheck'', ``validate'', ``re-evaluate''\}. 

Figure~\ref{fig:enhanced_verification} presents the results, where the proportion of responses that contain an explicit verification phrase consistently rises after \method{} training.  The increase is modest for the 1.5 B model (+0.05 absolute) but becomes substantial as scale grows (+1.09 for 3 B and +1.05 for 7 B).  Because both systems share the same decoding hyper-parameters, the higher frequency stems from the policy update that jointly optimizes for solution reward and verification reward, validating our first contribution: \method{} integrates verification into the RL loop instead of treating it as a post-hoc heuristic.

Simply asking the model to ``verify'' is not sufficient; the verification needs to be correct.   
\method{} yields clear improvements over Zero-RL for self-verified reasoning accuracy at every scale: +3.3\% (1.5 B), +1.7\% (3 B), and a striking +6.5\% (7 B).  These gains show that \method's online verifier reward shapes the policy toward not only producing more verifications, but also ones that align with ground truth.

The case in Table \ref{tab:verfication_case} illustrates this distinction.  Zero-RL ``verifies'' 222 by merely restating divisibility rules, overlooking the necessity of a trailing zero for multiples of 30.  \method, in contrast, recomputes both the units-digit constraint and the digit-sum divisibility test, ultimately validating the answer of 2220.  Such structured, multi-step verification reflects a genuinely internalized skill and explains the quantitative trend that higher verification frequency correlates with higher reasoning accuracy.

\begin{table}[htpb]
    \centering
    \caption{Examples of verification behaviors of different methods on Qwen2.5-7B: \method{} can incur more effectiveness self-verification in the reasoning.}
    % \resizebox{\linewidth}{!}
    {
    \begin{tabular}{p{0.1\linewidth} p{0.8\linewidth}}
    \toprule
    Problem & What is the least positive integer multiple of 30 that can be written with only the digits 0 and 2? \\
    \midrule
    Zero-RL & To find the least positive integer multiple of 30 that can be written with only the digits 0 and 2, we need to consider the properties of multiples of 30 \dots Since 222 is divisible by both 2 and 3, it is a multiple of 30.
{\bf Let's verify:} {\color{red} 222 is even, so it is divisible by 2. The sum of the digits of 222 is \(2 + 2 + 2 = 6\), which is divisible by 3.
Thus, 222 is indeed a multiple of 30.}
Therefore, the least positive integer multiple of 30 that can be written with only the digits 0 and 2 is \(\boxed{222}\).\\
    \midrule
    \method &   To find the least positive integer multiple of 30 that can be written with only the digits 0 and 2, we need to \dots Therefore, the smallest number \( N \) is \( 2220 \).
    {\bf To verify:} {\color{ngreen} \( 2220 \) ends in 0, so it is a multiple of 10. The sum of the digits of \( 2220 \) is \( 2 + 2 + 2 + 0 = 6 \), which is a multiple of 3. Since \( 2220 \) satisfies both conditions, it is the least positive integer multiple of 30 that can be written with only the digits 0 and 2.}
    Thus, the final answer is \(\boxed{2220}\).\\
    \bottomrule
    \end{tabular}}
    \label{tab:verfication_case}
\end{table}

\section{Conclusion}

In this work, we introduced \method, a novel online reinforcement learning framework that integrates problem-solving with explicit self-verification training for LLMs. By leveraging verifiable rewards for both generation and verification tasks within a unified RL objective, \method{} aims to overcome superficial self-reflection and foster more robust reasoning capabilities. Our experiments, primarily using PPO with Qwen2.5 models on diverse mathematical reasoning benchmarks, demonstrate that \method{} significantly improves problem-solving accuracy while concurrently developing strong self-verification skills. Notably, \method{} models learned to verify their own on-policy generations more effectively than off-the-shelf verifiers. Together, \method{} provides a promising direction for building more reliable and self-aware LLM reasoners, adaptable to various policy-gradient algorithms and extendable to other domains with verifiable rewards.

Future work includes: (1) exploring other challenging reasoning domains beyond mathematical reasoning, such as code generation, physics reasoning, and real-world
reasoning-intensive domains~\citep{su2025crossingrewardbridgeexpanding}; (2) examining additional policy-gradient algorithms besides PPO (such as GRPO~\citep{deepseek-math} and RLOO~\citep{ahmadian2024back}) within the \method{} framework; and (3) investigating potential synergies between \method{} and complementary enhancements like retrieval-augmented generation (RAG) or external tools for knowledge verification.

\bibliography{main}
\bibliographystyle{abbrvnat}

\newpage
\appendix
\section{Limitations}
\label{apx:limitation}
On experiments: After training the LLM to self-verify its outputs, it may serve as a pseudo–rule-based verifier to further guide RL training without relying on ground-truth labels. This opens the possibility of self-improvement on unlabeled data, which we leave for future exploration as it is beyond the current scope. Our experiments focus exclusively on math reasoning tasks across diverse problem types. While the generalization of the method to other domains remains underexplored, we argue that its strong performance on math reasoning is a compelling demonstration of its effectiveness. \method{} is expected to generalize as long as a well-defined verifiable reward is accessible.

On algorithm: \method{} trains the LLM as a generative verifier that produces natural language critiques, which has shown benefits for both reasoning and test-time scaling. An alternative design is to train a discriminative verifier with a separate classification head. While it remains unclear how \method{} would perform in that setting, we believe this does not affect our main contributions which demonstrate the effectiveness of generative verification in improving problem-solving capabilities.

\section{Prompt Templates}
\label{apx:prompt_templates}
\begin{figure}[h]
\centering
\begin{prompttext}
Below you are presented with a question and a tentative response. Your task is to evaluate and assign a rating to the response based on the following clear criteria:

\vspace{10pt} 
Rating Criteria:

\vspace{10pt}
1. Missing final answer enclosed in \textbackslash \textbackslash boxed\{\} at the end: assign \textbackslash \textbackslash boxed\{-1\}.

2. Correct response with the final answer enclosed in \textbackslash \textbackslash boxed\{\} at the end: assign \textbackslash \textbackslash boxed\{1\}.

3. Incorrect response with the final answer enclosed in \textbackslash \textbackslash boxed{} at the end: assign \textbackslash \textbackslash boxed\{-0.5\}.

\vspace{10pt} 
\#\#\# Question Begin \#\#\#

\{\textbf{Question}\}

\#\#\# Question End \#\#\# 

\vspace{10pt} 
\#\#\# Response Begin \#\#\#

\{\textbf{Response}\}

\#\#\# Response End \#\#\#

\vspace{10pt} 
Briefly summarize your analysis, then clearly state your final rating value enclosed in \textbackslash \textbackslash boxed\{\} at the end.
\end{prompttext}
\caption{Verification prompt used in the experiment.}
\label{fig:ver_prompt}
\end{figure}

\begin{figure}[h]
\centering
\begin{promptspecial}
<|im\_start|>system \\
Please reason step by step, and put your final answer within \textbackslash \textbackslash boxed\{\}.<|im\_end|> \\
<|im\_start|>user \\
\{\textbf{Input}\}<|im\_end|> \\
<|im\_start|>assistant
\end{promptspecial}
\caption{Prompt template used in the training and evaluation.}
\label{fig:qwen_prompt}
\end{figure}

\begin{figure}[h]
\centering
\begin{promptspecial}
<|im\_start|>system \\
You are a helpful assistant.<|im\_end|> \\
<|im\_start|>user \\
\{\textbf{Input}\} Please reason step by step, and put your final answer within \textbackslash \textbackslash boxed\{\}.<|im\_end|> \\
<|im\_start|>assistant
\end{promptspecial}
\caption{Prompt template used for Qwen base model evaluation.}
\label{fig:base_eval_prompt}
\end{figure}

\section{Training Details}
\label{apx:train_details} 
During RL training, we set the actor's clipping ratio to 0.2 and disable the KL penalty loss. The critic uses a clipping range of 0.5. The learning rates are fixed at $5 \times 10^{-7}$ for the actor and $9 \times 10^{-6}$ for the critic. The KL divergence coefficient is set to $1 \times 10^{-2}$. We limit the maximum response length to 3000 tokens, which already results in a negligible clip ratio. The full dataset is trained for 12 epochs. This configuration is shared across both the Zero-RL and \method{} models.

For the SFT baseline models, we use a batch size of 32 and apply a cosine learning rate scheduler with a learning rate of $2 \times 10^{-5}$ and a warm-up ratio of $1 \times 10^{-3}$. The dataset is trained for 3 epochs.

\section{Evaluation Details}
\label{apx:evaluation_details}
\subsection{Verification Evaluation with Other Verifiers}
\label{subsec:ver_eval}
To evaluate the verification accuracy of \method{} and GPT-4o (prompted as a verifier), we extract the final verification score from each response and normalize it to either +1 (predicted correct) or 0 (predicted incorrect). The normalization is defined as:
\begin{align*}
s_\text{normalized} &= \begin{cases}1, & s =1  \\ 0, & \text{otherwise} \end{cases} 
\end{align*}
which aligns with the criteria used by the rule-based outcome verifier. For the Math-Shepherd model, which outputs a continuous score in the range $[0,1]$ (with 0 indicating the solution/step is predicted to be incorrect and 1 indicating correct), we apply a threshold of 0.5 for normalization:
\begin{align*}
s_\text{normalized} &= \begin{cases}1, & s > 0.5  \\ 0, & \text{otherwise} \end{cases} 
\end{align*}

After normalization, we compute verification accuracy by directly comparing the predicted scores against those returned by the outcome verifier.

\subsection{Weighted Majority Voting with Self-Verification}
In \S~\ref{sec:tts}, we explore the combination of self-consistency and self-verification in test time following \citep{Wang2023MathShepherdVA}. In practice, we initially classify solutions into distinct groups according to their final answers. Following that, we extract and normalize the self-verification scores and normalize them as +1 (correct) and 0 (incorrect) as in \ref{subsec:ver_eval}. Since our score are binary and could lead to an unexpected zero sum, we integrate Laplace smoothing for computing the mean score for the answer.
Formally, the final selected answer based on $N$ candidate solutions is:
\begin{equation}
    a_{\text{maj@N+self-verify}} = \text{argmax}_{a} 
    \underbrace{\sum_{i=1}^N \mathbb{I}(a_i = a)}_{\text{frequency}} 
    \cdot 
    \underbrace{\frac{\alpha + \sum_{i=1}^N S(p, s_i)}{N + \alpha d}\vphantom{\sum_{i=1}^N}}_{\text{smoothed mean score}}.
\end{equation}
where $S(p, S_i)$ is the score of the $i$-th solution assigned by the self-verification. In practice, we set $\alpha=2$ and $d=2$ empirically, suggesting a prior belief of a 0.5 average score.

\section{Detailed Experiment Results}
\subsection{Detailed Comparison with off-the-shelf verifiers}
In \S~\ref{para:ver_compare}, we report the average verification accuracy across the five benchmarks. Here, we present the detailed verification accuracy comparison between \method{} models, Math-Shepherd, and GPT-4o on each evaluation benchmark.
\begin{figure}[h]
  \centering
  \includegraphics[width=.6\linewidth]{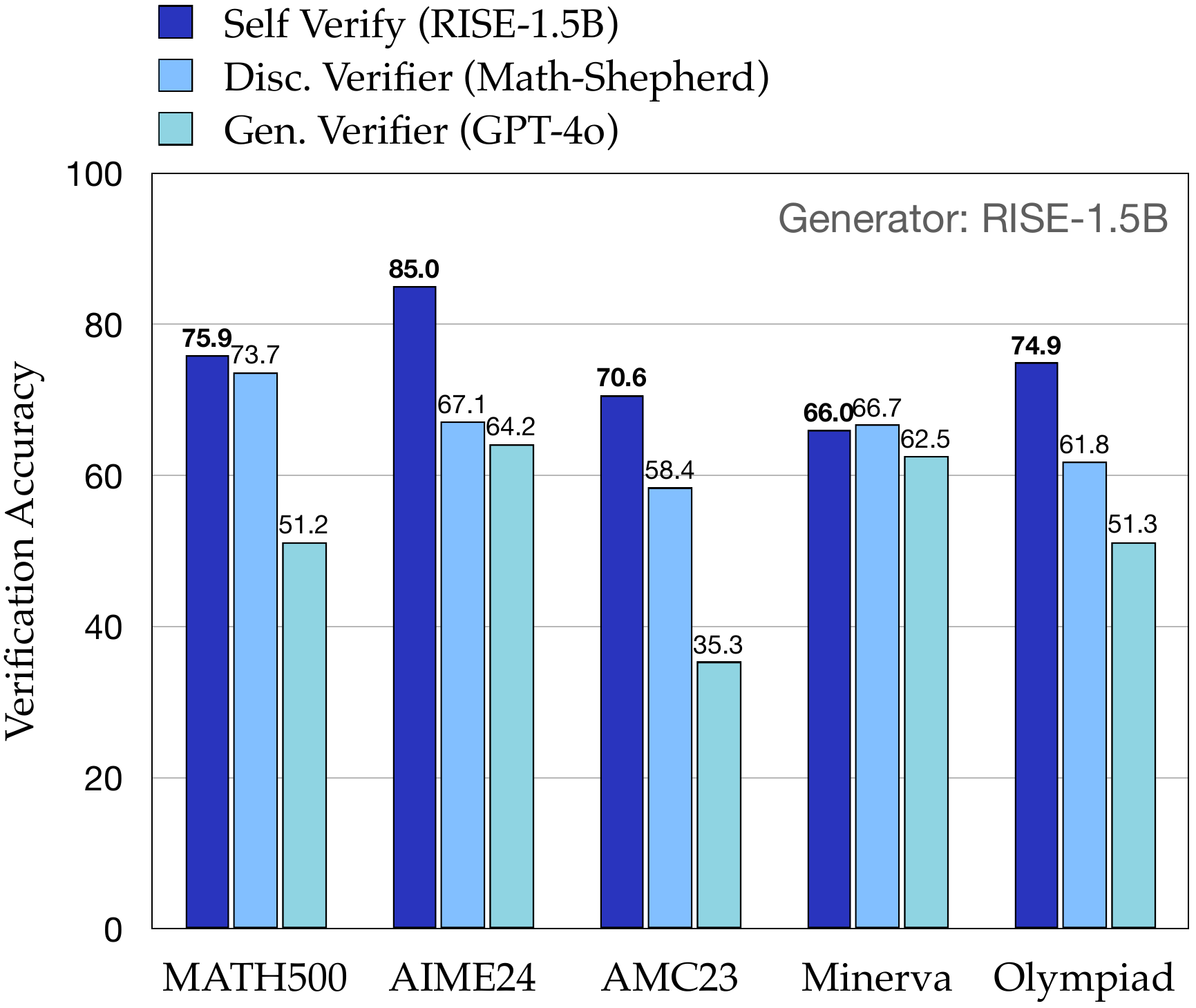}
  \caption{Detailed comparisons of verification accuracy between \method{}-1.5B and other verifiers.}
  \label{fig:v_acc_1.5b}
\end{figure}

\begin{figure}[h]
  \centering
  \includegraphics[width=.6\linewidth]{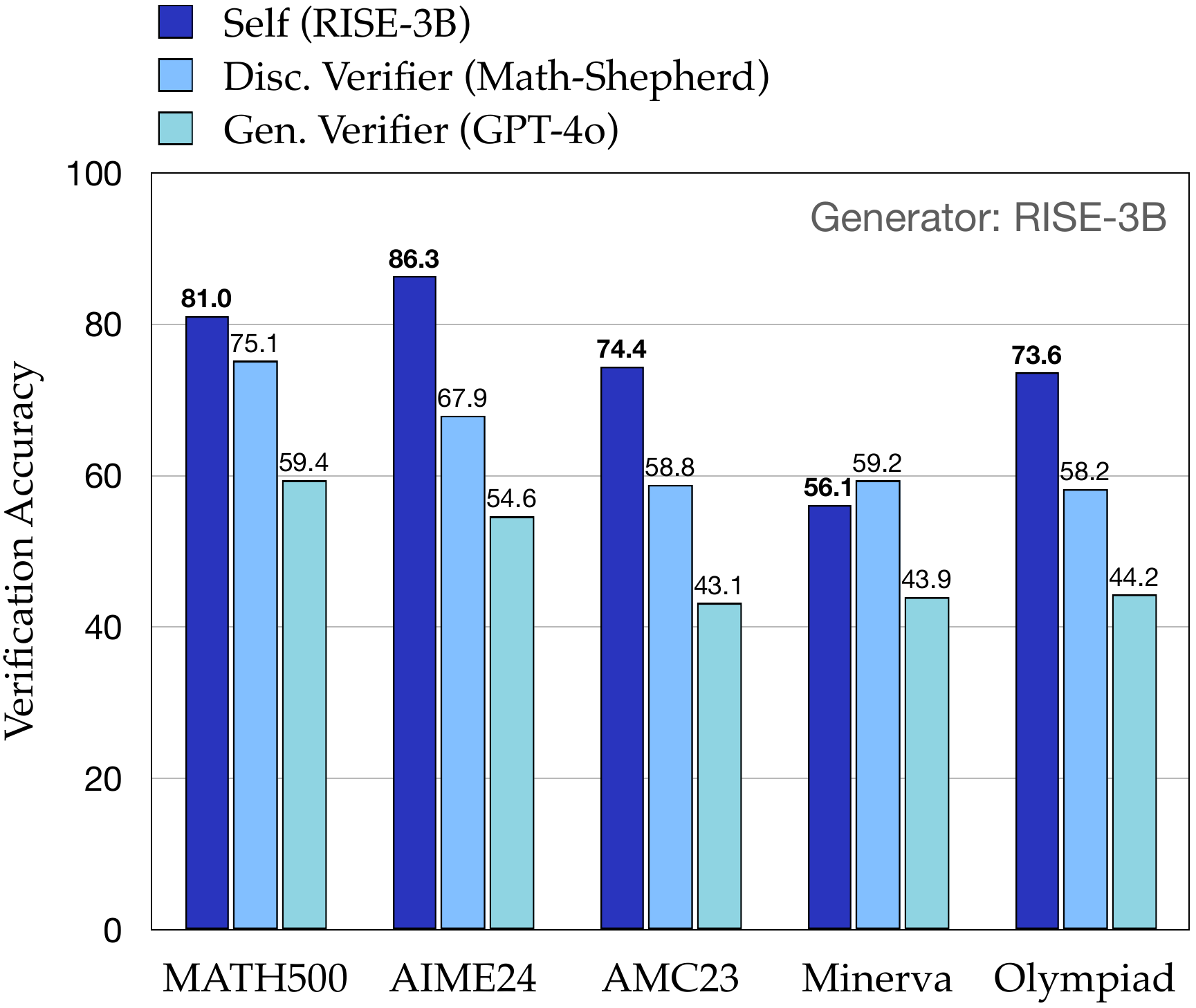}
  \caption{Detailed comparisons of verification accuracy between \method{}-3B and other verifiers.}
  \label{fig:v_acc_3b}
\end{figure}

\begin{figure}[h]
  \centering
  \includegraphics[width=.6\linewidth]{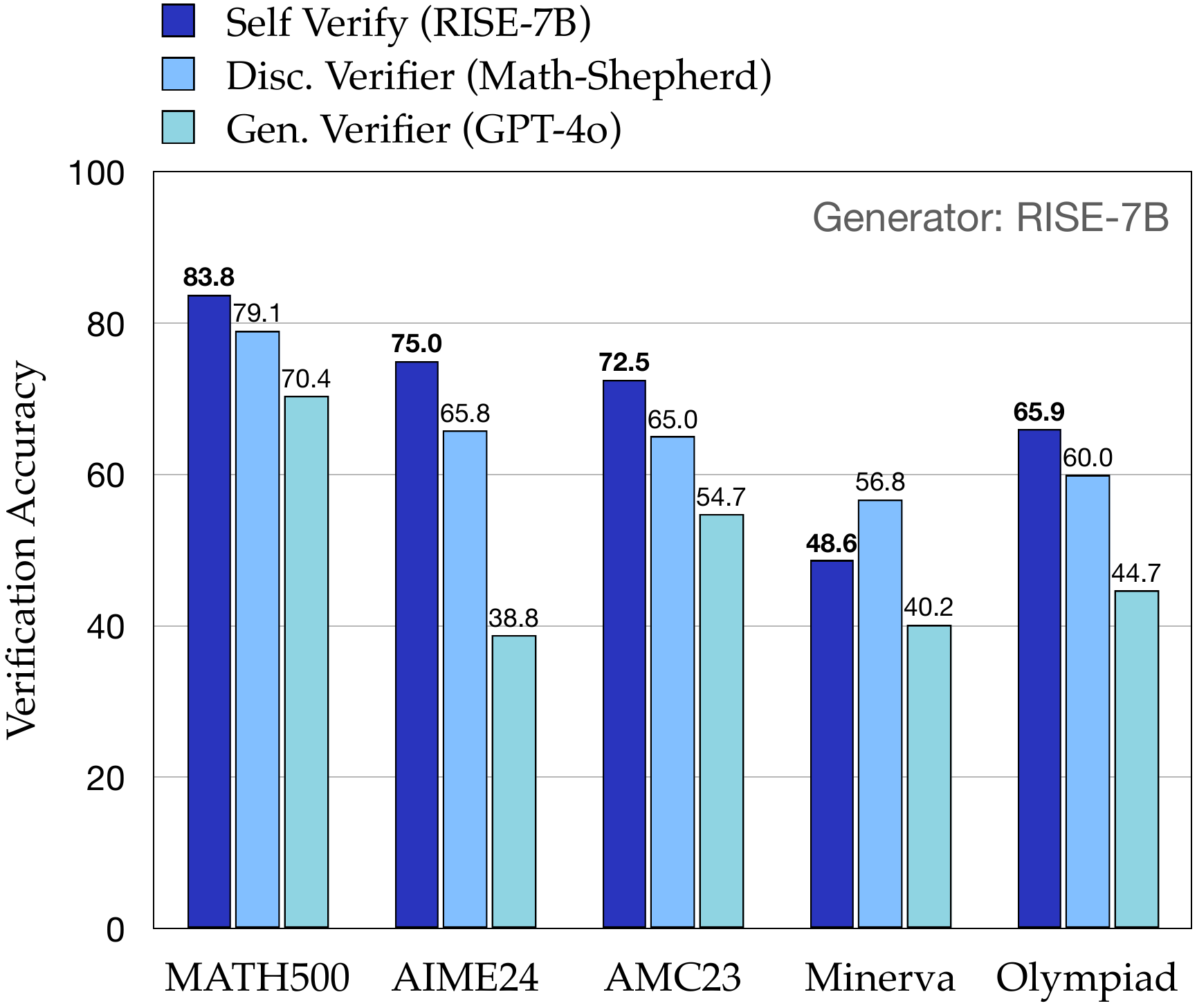}
  \caption{Detailed comparisons of verification accuracy between \method{}-7B and other verifiers.}
  \label{fig:v_acc_7b}
\end{figure}

\subsection{Detailed Analysis for Enhanced Verification}
In Figure~\ref{fig:enhanced_verification}, we report the average verification frequency and accuracy of self-verified solutions on the five benchmarks. Here, we present the fine-grained results between \method{} models and Zero-RL baseline on each evaluation benchmark.
\begin{table}[ht]
\centering
\caption{Performance comparison between \method{} models and Zero-RL models on verification frequency and effectiveness for the generation.}
\label{tab:reflection}
\resizebox{0.8\textwidth}{!}{
\begin{tabular}{lrrrrrr}
\toprule
\textbf{Method} & \multicolumn{6}{c}{\textit{Verification Frequency}} \\
\cmidrule(lr){2-7}
 & \textbf{MATH} & \textbf{AIME} & \textbf{AMC} & \textbf{Minerva} & \textbf{Olympiad} & \textbf{Avg.} \\ 
\midrule
Qwen2.5-1.5B-Zero-RL & 6.45 & 6.67 & \textbf{7.81} & 2.25 & \textbf{15.59} & 7.75 \\
\textbf{\method{}-1.5B} & \textbf{7.10} & \textbf{8.75} & 5.31 & \textbf{2.53} & 15.31 & \textbf{7.80} \\
\midrule
Qwen2.5-3B-Zero-RL & \textbf{4.90} & 8.33 & 14.29 & 2.99 & 8.72 & 7.85 \\
\textbf{\method{}-3B} & 4.63 & \textbf{9.17} & \textbf{18.18} & \textbf{3.08} & \textbf{9.67} & \textbf{8.94} \\
\midrule
Qwen2.5-7B-Zero-RL & 5.30 & 5.00 & 7.19 & 1.56 & 8.19 & 5.45 \\
\textbf{\method{}-7B}  & \textbf{6.08} & \textbf{7.92} & \textbf{8.13} & \textbf{1.79} & \textbf{8.57} & \textbf{6.50} \\
\bottomrule
\toprule
& \multicolumn{6}{c}{\textit{Self-Verified Solution Accuracy}} \\
\cmidrule(lr){2-7}
Qwen2.5-1.5B-Zero-RL & 37.21 & 0.00 & 24.00 & \textbf{24.49} & \textbf{15.59} & 20.26 \\
\textbf{\method{}-1.5B} & \textbf{38.73} & \textbf{4.76} & \textbf{35.29} & 23.64 & 15.31 & \textbf{23.55} \\
\midrule
Qwen2.5-3B-Zero-RL & \textbf{45.92} & 0.00 & 14.29 & 20.00 & 19.96 & 20.03 \\
\textbf{\method{}-3B} & 43.78 & \textbf{4.55} & \textbf{18.18} & \textbf{22.39} & \textbf{19.54} & \textbf{21.69} \\
\midrule
Qwen2.5-7B-Zero-RL & 62.74 & 0.00 & 8.70 & \textbf{35.29} & 26.70 & 26.68 \\
\textbf{\method{}-7B} & \textbf{65.43} & \textbf{5.26} & \textbf{38.46} & 28.21 & \textbf{28.73} & \textbf{33.22} \\
\bottomrule
\end{tabular}
}
% \vspace{-10pt}
\end{table}

\subsection{Reflection Keywords Analysis}
Following~\cite{yeo2025demystifying}, we track the self-reflection keywords \{``wait'', ``however'',	``alternatively'', ``retry'', ``recheck''\} to quantitatively measure the general reflection behaviors beyond the self-verification among the model problem-solving responses. In practice, we sum the total word counts for the keyword set and normalize it by the dataset size. 

The results in Table~\ref{tab:general_ref} show that our \method{} model constantly have a higher level of reflection-related behaviors than the Zero-RL models, indicating the positive effect of self-verification training.

\begin{table}[h]
\centering
\caption{Reflection Keywords Rate between \method{} models and Zero-RL models.}
\label{tab:general_ref}
\resizebox{0.8\textwidth}{!}{
\begin{tabular}{lrrrrrr}
\toprule
\textbf{Method} & \multicolumn{6}{c}{\textit{Verification Frequency in Generation}} \\
\cmidrule(lr){2-7}
 & \textbf{MATH} & \textbf{AIME} & \textbf{AMC} & \textbf{Minerva} & \textbf{Olympiad} & \textbf{Avg.} \\ 
\midrule
Qwen2.5-1.5B-Zero-RL & 0.16 & 0.40 & 0.26 & \textbf{0.16} & 0.29 & 0.25 \\
\textbf{\method{}-1.5B} & \textbf{0.19} & \textbf{0.45} & \textbf{0.29} & \textbf{0.16} & \textbf{0.32} & \textbf{0.28} \\
\midrule
Qwen2.5-3B-Zero-RL & 0.14 & 0.40 & \textbf{0.24} & 0.11 & 0.27 & 0.23 \\
\textbf{\method{}-3B} & \textbf{0.16} & \textbf{0.45} & 0.20 & \textbf{0.13} & \textbf{0.29} & \textbf{0.25} \\
\midrule
Qwen2.5-7B-Zero-RL & 0.13 & 0.38 & 0.23 & 0.08 & 0.23 & 0.21 \\
\textbf{\method{}-7B} & \textbf{0.14 }& \textbf{0.50} & \textbf{0.29} & \textbf{0.10} & \textbf{0.27} & \textbf{0.26} \\
\bottomrule
\end{tabular}
}
% \vspace{-10pt}
\end{table}

\section{Licenses}
\label{apx:licenses}
\textbf{Datasets and Benchmarks.} The training dataset is derived from MATH (MIT License). We evaluate on five benchmarks: MATH 500 (MIT License), AIME 2024 (CC0: Public Domain), AMC 2023 (Apache License 2.0), Minerva Math (license not found), and Olympiad Bench (MIT License).

\textbf{Framework.} RL training is based on verl v0.2 (Apache-2.0 license), and SFT training is based on trl~\citep{vonwerra2022trl} v0.14.0 (Apache-2.0 license). Evaluation is performed using vllm framework~\citep{kwon2023vllm} v0.7.2 (Apache-2.0 License) and the script is based on OpenMathInstruct-2~\citep{toshniwal2024openmathinstruct2}.

\textbf{Models.} We train our models based on the Qwen2.5 series. Specifically, Qwen2.5-1.5B\footnote{\url{https://huggingface.co/Qwen/Qwen2.5-1.5B}} and Qwen2.5-7B\footnote{\url{https://huggingface.co/Qwen/Qwen2.5-7B}} are released under the Apache License 2.0, while Qwen2.5-3B\footnote{\url{https://huggingface.co/Qwen/Qwen2.5-3B}} is released under a custom Qwen Research license. We also compare against Math-Shepherd\footnote{\url{https://huggingface.co/peiyi9979/math-shepherd-mistral-7b-prm}} model (license not found), and GPT-4o (accessed via OpenAI API, governed by OpenAI Terms of Use\footnote{\url{https://openai.com/policies/row-terms-of-use/}}).

\newpage
\section{Cases Study}
% Put some reasoning cases here
\begin{figure}[h]
\centering
\begin{casestudy}
\#\#\# Analysis:

The response provided contains several errors and incorrect calculations. Here's a step-by-step breakdown of the issues:

\vspace{10pt} 
1. **Triangle Type Identification**: The response incorrectly identifies the triangle as a right triangle based on the Pythagorean theorem, but it does not correctly apply the theorem to confirm the right angle.\\
2. **Angle Bisector Theorem Application**: The Angle Bisector Theorem is applied incorrectly, leading to an incorrect ratio for \(BD\) and \(DC\).\\
3. **Length Calculation of \(AD\)**: The length of the angle bisector \(AD\) is calculated incorrectly using the formula for the length of the angle bisector in a right triangle.\\
4. **Power of a Point Theorem**: The application of the Power of a Point theorem is incorrect, leading to an incorrect value for \(AP\).\\

\vspace{10pt} 
\#\#\# Final Rating:

\textbackslash \textbackslash boxed\{-0.5\}
\end{casestudy}
\caption{Example verification returned by \method{}-7B on AIME 24.}
\label{fig:ver_case}
\vspace{-20pt} 

\end{figure}

\end{document}